# An Improved 3D Skeletons UP-Fall Dataset: Enhancing Data Quality for Efficient Impact Fall Detection

Tresor Y. Koffi[1], Youssef Mourchid[1], Mohammed Hindawi[2], Yohan Dupuis[3]

[1]CESI LINEACT Laboratory, UR 7527, Dijon, France;
[2]CESI LINEACT Laboratory, UR 7527, Lyon, France;
[3]CESI LINEACT Laboratory, UR 7527, Paris La Defense, France

## ABSTRACT

Detecting impact where an individual makes contact with the ground within a fall event is crucial in fall detection systems, particularly for elderly care where prompt intervention can prevent serious injuries. The UP-Fall dataset, a key resource in fall detection research, has proven valuable but suffers from limitations in data accuracy and comprehensiveness. These limitations cause confusion in distinguishing between non-impact events, such as sliding, and real falls with impact, where the person actually hits the ground. This confusion compromises the effectiveness of current fall detection systems. This study presents enhancements to the UP-Fall dataset aiming at improving it for impact fall detection by incorporating 3D skeleton data. Our preprocessing techniques ensure high data accuracy and comprehensiveness, enabling a more reliable impact fall detection. Extensive experiments were conducted using various machine learning and deep learning algorithms to benchmark the improved 3D skeletons dataset. The results demonstrate substantial improvements in the performance of fall detection models trained on the enhanced dataset. This contribution aims to enhance the safety and well-being of the elderly population at risk. To support further research and development of building more reliable impact fall detection systems, we have made the improved 3D skeletons UP-Fall dataset publicly available at this link https://zenodo.org/records/12773013.

## 1. INTRODUCTION

Falls represent a major health risk for the elderly population, often leading to serious injuries, loss of independence, and in severe cases, fatality. Recent studies indicate that approximately one in four adults aged 65 and older experiences a fall each year, with fall-related injuries being a leading cause of hospitalization and fatalities among this population [1][2]. Effective fall detection systems are essential to minimize fall-related injuries and provide timely assistance to individuals at risk. Moreover, an immediate response can prevent complications, minimizing the severity of injuries, and potentially saving lives. Hence, the ability to accurately detect impact when an individual makes contact with the ground within fall event is crucial for the effectiveness of these systems. However, the development and improvement of fall detection systems heavily rely on high-quality datasets. These datasets serve as essential resources for training, testing, and benchmarking detection algorithms. At the same time, obtaining such data can be extremely challenging due to privacy concerns and ethical considerations [3]. Hence, creating comprehensive and accurate fall datasets presents unique challenges. Many datasets such as UMAFall dataset [4], URFall dataset have been used for fall detection [5] but the UP-Fall dataset has emerged as a key resource in fall detection research, providing valuable multimodal data for algorithm development [6] [7]. However, it suffers from limitations in data accuracy and comprehensiveness. These limitations have a direct influence on the performance of current fall detection systems, particularly in their ability to accurately detect impact within fall events. This inconsistent can delay the fall detection process, increase the false alert and compromise the effectiveness of these systems in providing timely assistance. Moreover, due to limited medical resources, detecting the impact where individual hits the ground allows better allocation of resources. To address these limitations, we propose and implement a preprocessing approach for the UP-Fall dataset to accurately detect the impact within fall events using 3D skeletons data. The main contributions of this paper are as follows:

1. We propose and implement a preprocessing approach for the UP-Fall dataset to accurately detect impact within fall events using 3D skeleton data.

2. We conduct extensive experiments using various machine learning and deep learning algorithms to benchmark the improved 3D skeleton UP-Fall dataset.

3. We make the enhanced 3D skeleton UP-Fall dataset publicly available, along with our preprocessing methodology, to facilitate further research and development in impact fall detection systems.

The rest of the paper is organized as follows: In Section 2, we review the related works on fall detection systems, and previous studies using the UP-Fall dataset, highlighting the gaps in current research that our study aims to address. In Section 3 details our proposed approach, describing the preprocessing technique for the UP-Fall dataset. In Section 4, we present our experimental and results, describing our methodology, the machine learning and deep learning algorithms used for benchmarking, and providing a comprehensive analysis of our findings. Finally, the Section 5 concludes the paper with a summary of our key contributions, and suggests potential directions for future research.

## 2. RELATED WORKS

Advancements in computer vision, driven by graphs, statistical techniques, and deep learning, have greatly enhanced visual data processing, which is particularly beneficial for improving fall detection accuracy [22,23,24] . These approaches can be classified into three main approaches such as threshold-based approaches [8] [9], which analyzes sensor values and compares them to a predefined value. Machine learning based-approaches [10] [11] offer improved adaptability and performance compared to threshold-based methods. They can handle more complex fall scenarios and often provides better generalization across different individuals and environments. More recently, deep learning based approaches [12] [13] excel at automatically learning relevant features from raw data, potentially eliminating the need for manual feature engineering. They can capture complex temporal and spatial patterns in fall data, especially in diverse and challenging real-world scenarios. However, due to the challenges of obtaining fall datasets, primarily because of privacy concerns and ethical considerations in simulating falls with elderly subjects, the UP-Fall dataset has become a valuable resource in fall detection research [17]. Several studies have used this dataset, leveraging its multi-modal data to develop and evaluate fall detection systems. Authors in [18] employed the UP-Fall dataset to evaluate a fall detection system based on machine learning techniques. Their work demonstrated the dataset's utility in developing and testing fall detection algorithms, achieving high accuracy in fall classification. An interesting work [19] proposed a fall detection system using the UP-Fall dataset, combining machine learning with complex event processing. Their approach demonstrated how the dataset could be used to develop sophisticated fall detection systems. While these studies have made significant contributions to fall detection research using the UP-Fall dataset, our investigation has revealed two critical gaps that significantly influences the dataset's reliability for accurate fall and impact detection. Upon visual inspection, we discovered few inconsistencies in the dataset's structure and labeling. Specifically, the folder labeled "not fall" contains images of fall events. Moreover, the existing labelization applied to the actual dataset uses Signal Magnitude Vector (SMV)-based labelization, where a threshold value of 1g is set to detect falls. This method, while useful, has limitations in accurately distinguishing between non-impact events, such as sliding, and real falls with impact, where the person actually hits the ground [19]. This mislabeling poses a significant challenge for researchers, as it can affect the algorithm performance and leads to false fall detection. Additionally, as noted by [18] there is a critical issue with the human skeleton data used for fall detection. Their work revealed that some frames in the UP-Fall dataset contain not only the main subject but also other individuals present in the background. This multiple skeletons can cause confusion in detecting the main subject, potentially compromising the accuracy of fall detection algorithms that rely on skeletal data, as the algorithm may mistakenly focus on the additional persons instead of the main subject performing the fall scenario. These inconsistencies in data organization and skeleton representation highlight the need for dataset improvement, particularly for detecting impact within fall event. By improving the UP-Fall dataset, we aim at correcting the folder structure and labeling of fall and non-fall events, develop methods to accurately identify and isolate the main subject in frames with multiple individuals, ensure accurate identification and labeling of impact moments within fall events using only 3D skeleton Data, and improve the overall reliability and usability of the dataset for impact fall detection research.

# 3. MATERIAL AND METHOD

## 3.1. Dataset

The UP-Fall dataset is a comprehensive collection of data on falls and daily activities, obtained from 17 healthy young adults, comprising 9 males and 8 females aged between 18 and 24 years. This dataset was collected using a multimodal approach with 14 distinct devices, including wearable sensors and cameras. The wearable sensors, attached to various Body parts, capture essential raw data from 3-axis accelerometers, 3-axis gyroscopes, and ambient light sensors. Simultaneously, cameras record high-resolution of sequential images capturing both lateral and frontal views of the subjects. The experimental set up for the data collection is shown in the Figure 1. As summarized in Table 1, as well as specific falling scenarios. Hence, this dataset provides valuable multimodal data for developing and evaluating fall detection systems.

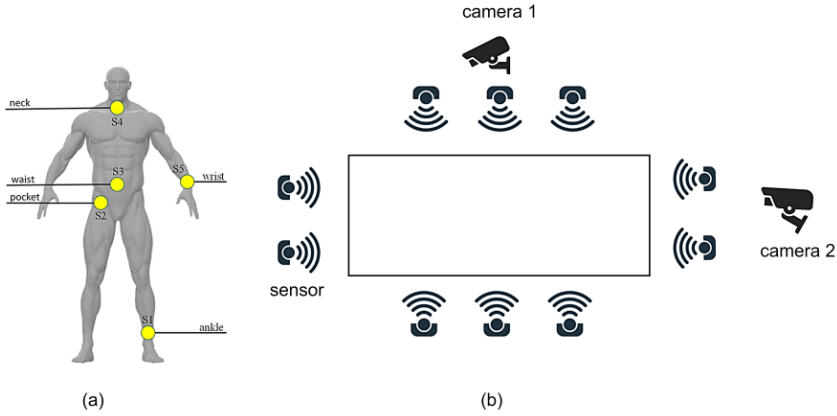

Figure 1. UP-Fall data acquisition. (a) Wearable sensors and EEG located at the human body. (b) Layout of the context-aware sensors and camera views position.

Table 1. Activities performed by subjects in UP-Fall data, adapted from [17]; ADL refers to Activity of Daily Life

| Activity ID | Category | Description | Duration (s) | Abbreviation |
|---|---|---|---|---|
| 1 | FALL | Falling forward using hands | 10 | FH |
| 2 | FALL | Falling forward using knees | 10 | FF |
| 3 | FALL | Falling backward | 10 | FB |
| 4 | FALL | Falling sideward | 10 | FS |
| 5 | FALL | Falling while attempting to sit in an empty chair | 10 | FE |
| 6 | ADL | Walking | 60 | W |
| 7 | ADL | Standing | 60 | S |
| 8 | ADL | Sitting | 60 | ST |
| 9 | ADL | Picking up an object | 10 | P |
| 10 | ADL | Jumping | 30 | J |
| 11 | ADL | Laying | 60 | L |

## 3.2. Data Pre-processing

Data pre-processing is a crucial step in our proposed approach to enhance the UP-Fall dataset. Effective pre-processing can improve the performance of impact fall detection systems by refining the raw data and addressing any inherent limitations. The multimodal nature of the UP-Fall dataset, which includes accelerometer, and image data, presents both opportunities and challenges that our pre-processing methodology aims to address.

### 3.2.1 Image Cropping and Background Removal

This preprocessing step starts by cropping the original images to 500 x 700 pixels to isolate the primary subject as shown in Figure 2b). This primary step aims to address the issue of multiple individuals in frames and reduce the noises in the frames that could occur during the pose estimation process and ensures that the extraction of the joints data are more accurate, as the algorithms can focus solely on the main subject. Additionally, we use the GrabCut algorithm [20] to remove the background elements and optimize the number frames detected by the pose estimation algorithm. Indeed, the UP-Fall dataset comprises sequences with varying numbers of frames per activity performed by each subject, with a maximum of 195 frames per sequence. Hence, without background removal, the pose estimation algorithm may detect fewer frames suitable for 3D joint extraction. By eliminating irrelevant background elements as shown in Figure 2c), we focus solely on the primary subject, significantly improving the detection of key points and joint data essential for precise impact fall detection. Consequently, this process contribute to a more reliable and consistent dataset for developing robust impact fall detection systems.

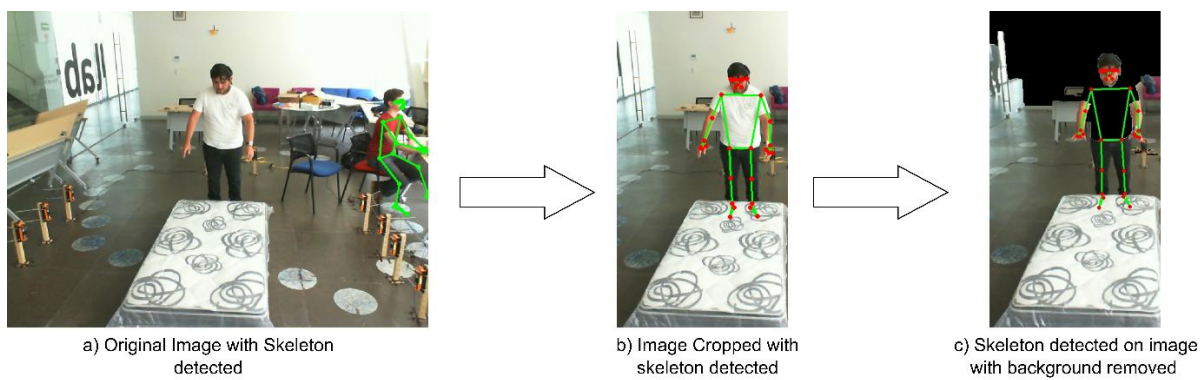

Figure 2. a) The pose estimation algorithm failed to detect joints from the main subject; (b) the cropped isolate the main subject and the pose estimation algorithm detect the joints; c) skeleton detected on image with the background removed

### 3.2.2 3D Joints Skeleton Data using Mediapipe BlazePose

After isolating the subject and cleaning the background, we extract 3D skeletons data using Mediapipe BlazePose as shown in Figure 3. This process detects and tracks up to 33 key points on the body, providing a detailed representation of human posture and movement during the fall events. BlazePose algorithm ensures accurate detection of the joints in complex fall scenarios, significantly improving the dataset's utility for developing impact detection models. This skeletal information enhances the dataset's comprehensiveness, enabling more precise impact detection and improving the overall reliability of impact fall detection systems.

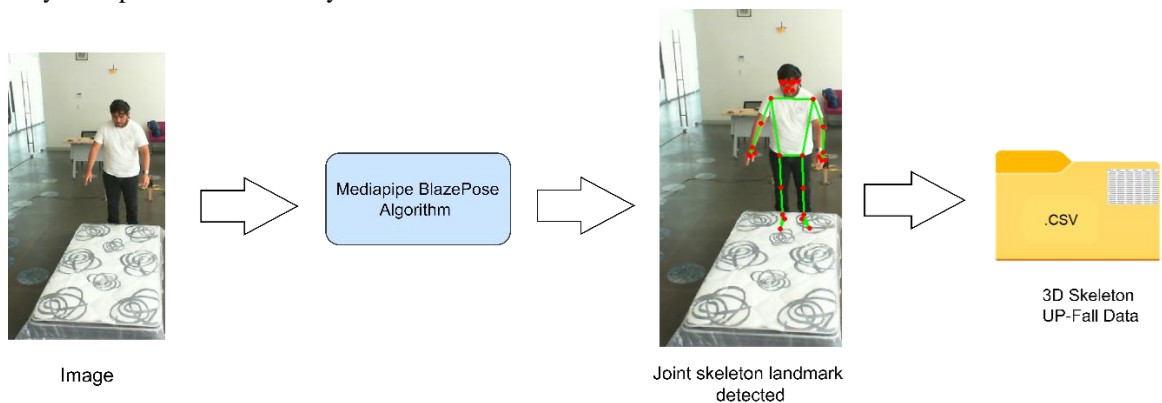

Figure 3. Extraction of the 3D skeletons Data using Mediapipe BlazePose Algorithm

### 3.2.3 Manual Labeling

Manual labeling is a critical step in preprocessing the UP-Fall dataset to address initial labelization issue existing. Our investigation revealed that some images in the "fall" folder do not actually exhibit impact and contain non-fall data as shown in Figure 4. This mislabeling could significantly compromise the performance of fall detection models leading to false alert. To improve this, we implemented a manual labeling process involving careful visual inspection and assign each frame to his corresponding folder. We therefore, review and categorize the fall events, ensuring that labels accurately represent the corresponding scenarios, with a specific focus on identifying true impact within the falls as shown in Figure5. While this process is labor-intensive and time-consuming, it is crucial for enhancing the dataset's reliability for impact detection.

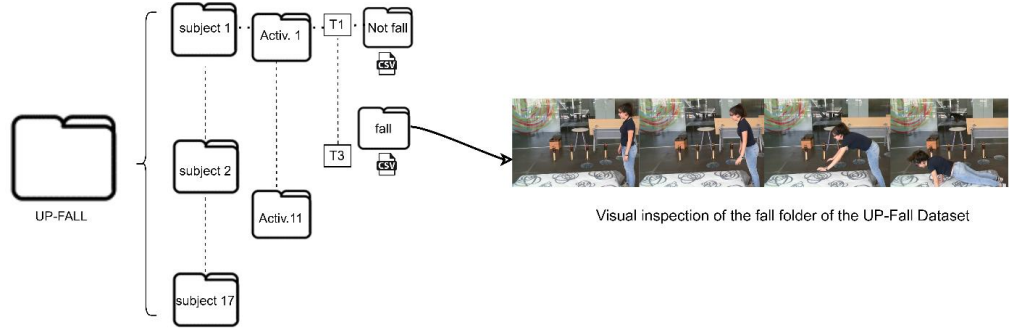

Figure 4. Example of the incoherence in the UP-Fall dataset

### 3.2.4 Semi-Automatic Labeling

This section introduces an enhancement to the UP-Fall dataset aimed at automatically detecting impact within fall events, addressing the time-consuming issue from the manual labeling. The dataset includes multimodal data from cameras and accelerometers. We focus on the inertial data by calculating the Signal Magnitude Vector (SMV) [21], which condenses information from the accelerometer's three axes (x, y, z) into a single value:

$$SMV = \sqrt{a_x^2 + a_y^2 + a_z^2} \qquad (1)$$

Where $a_x$, $a_y$ and $a_z$ represent the acceleration along x-axis, y-axis and z-axis respectively. To detect the impacts, we compare the SMV value against a predefined threshold:

$$\begin{cases} \text{if SMV} > \beta, \text{then Impact detected} \\ \text{if SMV} < \beta, \text{then no impact detected} \end{cases} \qquad (2)$$

Here, $\beta$ is the threshold value, set at $2g$. Impacts are detected when the SMV exceeds this threshold. Each detected event is then validated through visual inspection of the corresponding frame, ensuring accurate labeling of impact (1) and non-impact (0) events as shown in Figure5. This validation aligns detected events with visual data, identifying discrepancies and reducing false alerts where the algorithm accurately detected the impact within the fall events.

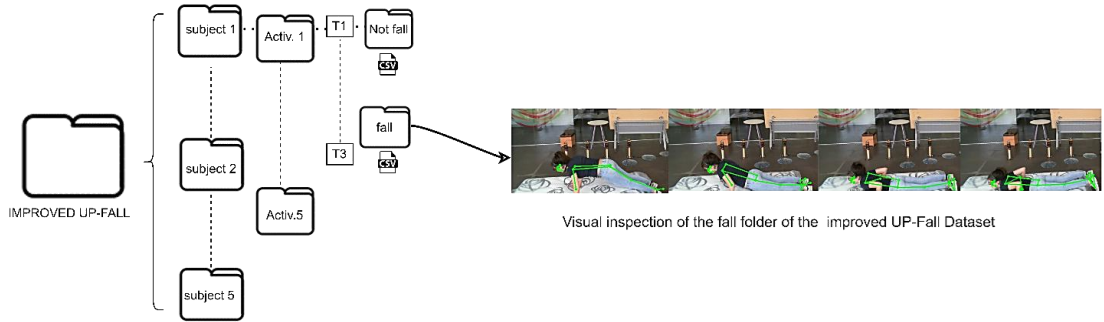

Figure 5. The Improved 3D skeletons UP-Fall Dataset for Impact Fall Detection

## 4. EXPERIMENTAL RESULTS

### 4.1. Data Improved 3D Skeleton UP-Fall Dataset

Our improved 3D skeletons UP-Fall dataset includes data from five (5) subjects performing five (5) fall scenarios including the activities of the daily life (ADLS). The dataset is designed to address limitations found in the original UP-Fall dataset and enhance its use for detecting impacts within fall events. We used Mediapipe pose estimation to extract joints from the body, providing 3D skeleton data. This process extracted up to 33 joints from each body within the frame, offering detailed 3D skeleton data for each subject during the fall scenarios.

### 4.2. Evaluation of Machine and Deep Learning Approaches

For our evaluation protocol, we divided our improved 3D skeletons dataset into three subsets: a training set (80%), a validation set (10%), and a test set (10%). To assess the performance of the models trained on the improved 3D skeletons UP-Fall dataset, we used several evaluation metrics such as accuracy, recall, precision, F1-score, sensitivity, and specificity.

#### 4.2.1 Evaluation Using Machine Learning Algorithms

To evaluate the effectiveness of the improved 3D skeletons UP-Fall dataset for impact detection in fall events, we employed various machine-learning algorithms, including Random Forest (RF), Support Vector Machine (SVM), Multi-Layer Perceptron (MLP), K-Nearest Neighbors (KNN), Naive Bayes (NB), and Stochastic Gradient Descent (SGD). The performance of each algorithm was assessed using the preprocessed dataset. All models demonstrated high performance, with accuracies and F1-scores above 98% as shown in Table 2. These results demonstrate that the improved 3D skeletons UP-Fall dataset significantly enhances the ability to detect impact within fall events, potentially leading to more accurate and reliable fall detection systems.

Table 2. Performance of machine learning Models using our improved 3D skeletons UP-Fall Dataset; The bold refers to the highest values

| Model | Dataset | Accuracy (%) | Precision (%) | Recall (%) | F1-Score (%) |
|---|---|---|---|---|---|
| RF | Improved Data | 99.28 | 98.47 | 99.18 | 98.47 |
| MLP | Improved Data | 99.47 | **99.70** | 97.83 | 98.74 |
| KNN | Improved Data | 98.35 | 98.45 | 99.03 | 98.74 |
| NB | Improved Data | 98.85 | 97.57 | 97,94 | 97.76 |
| SGD | Improved Data | 94.47 | 97.71 | 95.85 | 96.77 |
| SVM | Improved Data | **99.50** | 99.50 | **99.50** | **99.50** |

#### 4.2.2 Evaluation Using Deep Learning Algorithms

Following our machine learning evaluation, we assessed the improved 3D skeletons UP-Fall dataset using deep learning algorithms such as Convolutional Neural Network (CNN), Long Short-Term Memory (LSTM), Bidirectional

Long Short-Term Memory (BiLSTM) and graph based approaches [14,15,16] such as spatio-Temporal Graph Convolutional Networks (STGCN). These algorithms were specifically employed to detect impacts within fall events using the improved 3D skeletons data. As shown in Table3, the results show significant improvements in impact detection. The results demonstrate that our models perform well in detecting impact within fall events. Additionally, the LSTM appears to be the most effective model, achieving the highest average result among the evaluated algorithms. This makes the LSTM a promising model for real-world applications in impact fall detection.

Table 3. Performance of Deep Learning Models using Our Improved 3D skeletons UP-Fall Dataset; The bold refers to the highest values

| Model | Dataset | Accuracy (%) | Precision (%) | Sensitivity (%) | Specificity (%) |
|---|---|---|---|---|---|
| CNN | Improved Data | 93.00 | 95.27 | **100** | 78.79 |
| BiLSTM | Improved Data | 94.00 | 95.89 | **100** | 81.82 |
| STGCN | Improved Data | 92.00 | 91.43 | 95.59 | 84.38 |
| LSTM | Improved Data | **98.50** | **97.83** | 97.76 | **100** |

### 4.2.3 Visual Comparison between our Method and the state of the Art Fall Detection Algorithm

To provide a comprehensive evaluation of our improved 3D skeletons UP-Fall dataset, we conducted a visual comparison between our method and a state-of-the-art fall detection algorithm. This comparison serves to highlight the enhanced accuracy and reliability of our dataset in detecting impacts within fall events. As illustrated in Figure 6(a), upon visual inspection of the output predicted, the current algorithms using the original UP-Fall dataset failed to detect the actual impact where the person makes contact with the ground. In other hand our algorithms, using the improved 3D skeletons dataset accurately detects impact within fall event upon visual examination as shown in Figure 6(b). This contributes to the development of a more reliable impact fall detection system.

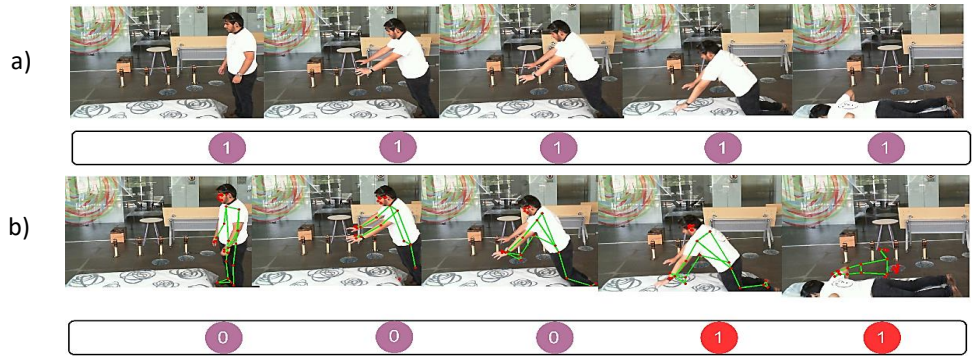

Figure 6. a) Trained on the current UP-Fall dataset, up on visual inspection the LSTM algorithm failed to accurately detect impacts; b) trained on the improved 3D skeletons dataset the LSTM accurately detect impacts upon visual inspection

## 5. CONCLUSION

In this study, we have successfully enhanced the UP-Fall dataset by introducing 3D skeletons data, addressing a critical limitation in fall detection research. This improvement provides a more detailed and accurate representation of human posture and movement during fall events, enabling more precise impact detection. Our methodology for extracting and integrating 3D skeletal information ensures the enhanced dataset's reliability. The importance of this improved 3D skeletons dataset for impact fall detection research and development is significant, especially given the need for effective impact fall prevention and response systems in aging populations. Future work could focus on expanding the dataset to include a more fall and subject particularly elderly individuals. Additionally, exploring advanced deep learning techniques to further refine the 3D skeletons data and developing algorithms for effective impact detection within fall events.